%% file: gestos.tex
\documentclass{fcs}

\usepackage{bm}
\usepackage{verbatim} 
\usepackage{url} 
\usepackage[table]{xcolor}
\usepackage[normalem]{ulem}

\usepackage[top=2cm,left=2cm,right=2cm]{geometry}

\volumn{ }
\doi{ }
\articletype{RESEARCH~ARTICLE}
\copynote{{\copyright} PREPRINT}
\ratime{}
\email{fronchetti at lidi.info.unlp.edu.ar}
\title{$\bm{Frontiers~ of~ Computer~ Science}$\\[2mm] Distribution of Action Movements (DAM): \\A Descriptor for Human Action Recognition}
\author{Franco RONCHETTI \xff, Facundo QUIROGA, Laura LANZARINI, Cesar ESTREBOU}
\address{{ Instituto de Investigacion en Informatica III-LIDI, Facultad de Informatica, Universidad Nacional de La Plata, Argentina}}

\begin{document}
\maketitle
\setcounter{page}{1}
\setlength{\baselineskip}{14pt}

\input{tex/notation}

\begin{abstract}

\input{tex/abstract}
\end{abstract}


\section{Introduction}
\input{tex/intro}

\section{Descriptor}
\label{descriptor}
\input{tex/descriptor}

\section{Experiments}
\label{experiments}
\input{tex/experiments}

\section{Conclusion}
\label{conclusion}
\input{tex/conclusion}

%\begin{thebibliography}{99}
%\input{tex/bibliography}
%\end{thebibliography}

\bibliographystyle{unsrt}
\bibliography{tex/ref}

\input{tex/biographies}

\end{document}

%% file: tex/notation.tex
% !TeX spellcheck = en_US

%set of actions
\newcommand{\as}{\mathcal{A}}

% action positions without subscript i
\newcommand{\ac}{\mathbf{A}}
\newcommand{\afj}{\mathbf{A}_{f,j}}
\newcommand{\aj}{\mathbf{A}_j}
\newcommand{\af}{\mathbf{A}_f}
% action positions with subscript i
\newcommand{\ai}{\mathbf{A}^i}
\newcommand{\aifj}{\mathbf{A}_{f,j}^i}
\newcommand{\aifjp}{\mathbf{A}_{f+1,j}^i}
\newcommand{\aifp}{\mathbf{A}_{f+1}^i}
\newcommand{\aij}{\mathbf{A}_j^i}
\newcommand{\aif}{\mathbf{A}_f^i}

% frames of action i
\newcommand{\fri}{F_i}

% action directions without subscript i
\newcommand{\dc}{\mathbf{D}}
\newcommand{\dcj}{\mathbf{D}_j}
\newcommand{\df}{\mathbf{D}_f}
\newcommand{\dfj}{\mathbf{D}_{f,j}}

% action directions with subscript i
\newcommand{\di}{\mathbf{D}^i}
\newcommand{\dij}{\mathbf{D}_j^i}
\newcommand{\dif}{\mathbf{D}_f^i}
\newcommand{\difp}{\mathbf{D}_{f+1}^i}
\newcommand{\difw}{\mathbf{D}_{f+(w-1)}^i}
\newcommand{\difj}{\mathbf{D}_{f,j}^i}

% action window without subscript i
\newcommand{\wc}{\mathbf{W}}
\newcommand{\wcj}{\mathbf{W}_j}
\newcommand{\wf}{\mathbf{W}_f}
\newcommand{\wfj}{\mathbf{W}_{f,j}}

% action winwow with subscript i
\newcommand{\wi}{\mathbf{W}^i}
\newcommand{\wij}{\mathbf{W}_j^i}
\newcommand{\wif}{\mathbf{W}_f^i}
\newcommand{\wifj}{\mathbf{W}_{f,j}^i}

\newcommand{\wall}{\mathcal{W}^{all}}

\newcommand{\pca}{P(c | \ac)}
\newcommand{\pcaf}{P(c | \af)}

\newcommand{\pcw}{P( c | \wc)}
\newcommand{\pcwf}{P( c | \wf)}

\newcommand{\pccl}{P( c | \cl)}

\newcommand{\cw}{\mathcal{C}_w}
\newcommand{\ci}[1]{\mathbf{C}^{#1}}
\newcommand{\cl}{\ci{l}}

\newcommand{\fw}{F^W}

% Histogram

\newcommand{\hc}{\mathbf{H}}
\newcommand{\hl}{H_l}

% UTILITY COMMANDS

% captions that wrap text
\newcommand{\wrappingcaption}[1]{
\caption[asd]{\begin{minipage}[t]{.8\linewidth} 
#1 
\end{minipage}
}
}

%% file: tex/abstract.tex
% !TeX spellcheck = en_US

%Human action recognition is a challenging computer vision and machine learning problem with several important applications, such as human computer interaction, surveillance, home monitoring, etc. In this paper, we present a novel human action recognition method using 3D skeletal information. The method adapts ProbSom's \cite{Estrebou2011} approach of dealing with temporal data by representing an action with the distribution of the directions of the movements contained in it. While the descriptor is global in the sense that it represents the overall distribution of movement directions, it is able to retain some temporal structure by concatenating information from several frame directions into a single frame.
%
%The descriptor, together with a standard classifier, outperforms several state of the art techniques on many well-known datasets.
%
%\vspace{10pt}
%Another version: (ELEGIR!!)
%\vspace{10pt}

Human action recognition from skeletal data is an important and active area of research in which the state of the art has not yet achieved near-perfect accuracy on many well-known datasets. In this paper, we introduce the Distribution of Action Movements Descriptor, a novel action descriptor based on the distribution of the directions of the motions of the joints between frames, over the set of all possible motions in the dataset. The descriptor is computed as a normalized histogram over a set of representative directions of the joints, which are in turn obtained via clustering. While the descriptor is global in the sense that it represents the overall distribution of movement directions of an action, it is able to partially retain its temporal structure by applying a windowing scheme.

The descriptor, together with a standard classifier, outperforms several state-of-the-art techniques on many well-known datasets.

%% file: tex/intro.tex
% !TeX spellcheck = en_US

\subsection{Motivation}

\noindent Human Action Recognition remains an unsolved problem in the fields of computer vision and pattern recognition.
It can be employed to monitor patients, detect suspicious activities, classify work situations as dangerous or safe, recognize player actions in games, or to provide touch-less human computer interactions.
Traditionally, vision-based human action recognition techniques used low-level image features as input to action models. In recent years, researchers have been increasingly using Mocap systems or sensors such as the Microsoft Kinect to obtain skeletal information of users performing actions. This trend has been motivated by the low cost and wide availability of this types of capture devices, and growing evidence supporting the hypothesis that skeletal data provides a superior representation for human action recognition \cite{yao2011does}.

\subsection{Overview}

This paper presents the Distribution Action Movements Descriptor (DAM), a novel descriptor for human actions based on the distribution of the movements of each joint in a skeletal representation of the user. 

We extend and adapt ProbSom's approach to person identification by voice, described in \cite{Estrebou2010}, to the case of human action recognition using skeletal information. The following is an outline of the computation of the resulting descriptor, explained in full in section \ref{descriptor}. 

We have a set of action recordings performed by persons, which are composed of a variable number of frames, where for each frame we know the 3D position of a given number of joints of the body. We first transform each action, replacing the absolute 3D position of each joint by the direction of its motion between frames. Thus, we obtain $n-1$ \textit{direction frames} from the original $n$ \textit{absolute position frames} of the action. Note that the direction frames are translation invariant. Previously, each joint of each action was smoothed and resampled individually; the first to remove spurious directions, the second to ensure each action has the same number of frames. The resulting directions of motion are 3D vectors, one for each frame and joint. We apply a windowing scheme to preserve temporal information, generating windowed direction frames (WDFs) from the direction frames. Then, we perform a clustering of the set of all the WDFs of the set $\ac$ of all actions to obtain a reduced set $\cw$ of direction frames that efficiently represents the distribution of directions of $\ac$. 

The descriptor of a new action $\ac$ is then computed as the normalized histogram of its direction frames over the set $\cw$, where each direction frame in  $\cw$ is a bin, and each direction frame of the action is assigned to the bins according to the same distance metric employed in the generation of  $\cw$.

This novel descriptor achieves excellent results on two well known datasets, MSRC12 and MSR Action3D, described in section \ref{experiments}. To the best of our knowledge, it surpasses the classification accuracy of almost all other state-of-the-art techniques on standard datasets to date (see also section \ref{experiments}).

%The paper is organized as follows. Section \ref{related} describes previous approaches on the problem of action recognition with skeletal information.  Section \ref{descriptor} provides a detailed explanation of the descriptor and the classification scheme developed. Sections \ref{databases} and \ref{experiments} describe the databases and experiments done, and finally section \ref{conclusion} includes final remarks and conclusion.

\subsection{Related work}
\label{related}

As described in \cite{hussein2013human}, there are three main challenges in a Human Action Recognition System: data capture, feature descriptors and action modeling. Our method does not deal with the first challenge, assuming the 3D skeletal data is somehow available, captured for example by employing a Mocap device or the Kinect as is the case with the datasets used in our experiments. The second and third challenges are very much intertwined, and are, in general, solved jointly.

Feature descriptors can be global, in the sense that they represent the action as a whole, or they can be local, in the sense that a separate descriptor is computed for each frame or interest point of the action. In the latter case, additional steps are needed to create an effective descriptor, for example, by modeling the temporal dependence with a Hidden Markov Model, or aggregating each local descriptor to generate a global one.

In recent years, there has been much interest in finding appropriate descriptors for action recognition.  

In \cite{barnachon2014ongoing}, Barnachon et al. apply a clustering algorithm to all the poses of every action sample, and then define a pose-based integral histogram for actions, computed over the resulting clusters, as descriptor. Actions can thus be encoded with this representation, and then compared using the Bhattacharyya distance. Since the descriptor clusters the poses directly, it is not translation-invariant.

Hussein et al. \cite{hussein2013human} utilize a covariance descriptor for each action, which is computed as the covariance matrix of the set of poses that comprise an action and contains as information the variance of movement of each joint (diagonal elements) and the relative movement between joints (non-diagonal elements). The descriptor has the disadvantage of being high dimensional. To build a model, they compute the descriptor for all training actions and then train an off-the-shelf  support-vector machine (SVM) classifier with a linear kernel. 

The approach in Gowayyed et al. \cite{gowayyed2013histogram} called Histogram of Oriented Directions, is similar to ours, in that it employs a histogram of directions, but their approach calculates histograms independently for each joint, and uses, not the 3D positions of the joint themselves, but the three 2D projections to planes $xy$, $xz$ and $yz$ of their position in the virtual coordinate space. The set of bins is defined arbitrarily as a regularly spaced grid. By calculating histograms on each joint separately, their descriptors lacks synchronization information between the joints in multi-joint actions. As in the method of Hussein et al., the classification is performed by combining the descriptor with a SVM.

Since this three descriptors all lack temporal information, a sub-action decomposition or temporal pyramid scheme was applied to the descriptors to generate more complex ones that retain some of the sequence information of the original actions.

The sequence of most informative joints (SMIJ) representation introduced by Ofli et al. \cite{ofli2014sequence} divides each action into a number of segments, and for each segment computes a ranking of the most informative joints according to an entropy-based measure, and chooses the $k$ most relevant ones to describe the segment. These are concatenated into a descriptor, and standard SVM and k-nearest neighbours classifiers were trained using it as input. The representation completely disregards any trajectory information contained in the action, since it only measures the amount of activity in the joints.

In \cite{cho2013classifying}, Cho et al. train a hybrid Multilayer Perceptron/Deep auto-encoder on a set of features composed of the 3D direction vector between joints for each frame, rotated and scaled to canonical angles and sizes to obtain invariances.

Li et al. \cite{li2010action} do not employ a full pose estimation technique, but instead sample 3D points belonging to the person's body from depth maps. They quantize the set of all poses from the training samples via clustering, and build an action graph using the cluster centers or \textit{salient poses} as nodes, inferring the transition probabilities from the training samples as well. Afterwards, they apply a decoding scheme based on Bi-gram with Maximum Likelihood Decoding (BMLD) to classify new samples according to the graph information.

None of these techniques achieves perfect accuracy on cross-subject cross-validation experiments in the datasets used to test them.

%Robust 3D Action Recognition with Random
%Occupancy Patterns
%Jiang Wang 
%
%Mining Actionlet Ensemble for Action Recognition with Depth Cameras
%Jiang Wang
%
%
%Classifying and Visualizing Motion Capture Sequences
%using Deep Neural Networks
%Kyunghyun Cho
%
%Action Recognition Based on A Bag of 3D Points
%Wanqing Li
%
%
%Mining Actionlet Ensemble for Action Recognition with Depth Cameras
%Jiang Wang
%
%
%Ongoing human action recognition with motion capture
%Mathieu Barnachon
%
%

%

%% file: tex/descriptor.tex
% !TeX spellcheck = en_US

In this section we develop the formulation for the descriptor and classification model. We begin by introducing a notation for actions.

\subsection{Notation}
Let $\as=\{ \ac^1,\dots,\ac^N\}$ be a set of $N$ actions, $\fri$ the number of frames of action $\ac^i$ and $J$ the number of joints (the same for every action). 

Each action is composed of an ordered sequence of frames. A frame contains the position of each joint of the person in that frame. So we have:
\begin{enumerate}
\item $\ai$: $ith$ action, with $\fri$ frames, where each frame contains the position of the $J$ joints.
%\item $\aij$: the joint positions for all $\fri$ frames of action $i$.
\item $\aif$: the positions of the $J$ joints of action $i$ in frame $f$, ie, a \textit{position frame}.
\item $\aifj$: the position of joint $j$ of action $i$ in frame $f$.
\end{enumerate}

For example, the position of joint $1$ of action $3$ in frame $5$ is denoted as $\mathbf{A}_{5,1}^3$ and the position of \textit{all} joints in the same frame and action is $\mathbf{A}_{5}^3$. Each position $\aifj$ is a vector in $\mathbb{R}^3$.

The notation for Direction Frames and Windowed Direction Frames (to be defined shortly) we use is analogous to that of positions.

%Similarly, we can define      $\dc$, $\dci$, $\dcj$, $\dcf$ and $\dcfj$ as before, where we replace the word position for direction. 

\subsection{Preprocessing}

The preprocessing steps are outlined in Figure \ref{preprocessing_diagram}.

First, the joint positions of each action $\aifj$ are smoothed with a gaussian-distributed weighted moving average (WMA). While simple, this step is essential to the effectiveness of the descriptor, since otherwise sensing errors or short spurious movements distort long term trends in the direction of movement of the joints on which we rely for classification.

Afterwards, the same joint positions $\aifj$ are re-sampled to a constant number of frames $F$, which is the same for all actions and joints, so that at the end of this process $\fri=F$ for all $i$. The re-sampling is done at uniformly spaced points along the arc-length of the path traced by the joint through all the frames of the action. The position of the joints in the new frames are estimated by cubic splines interpolation. 

Since each joint is re-sampled individually and with an arc-length parametrization, this step could distort the temporal synchronization between joints, but visual inspection in the datasets used for the experiments revealed no noticeable warping.

To obtain a translation invariant descriptor, we calculate for each joint and frame a direction vector $\difj \in \mathbb{R}^3, f=1,\dots,F-1$, that represents the direction vector between positions $\aifjp$ and $\aifj$, such that:

\begin{equation}
\label{delta} 
\difj = \aifjp - \aifj, \quad f=1,\dots,F-1
\end{equation}

By analogy, we call each $\dif$ a \textit{direction frame}, since it contains the direction of motion of each joint between the position frames $\aifp$ and $\aif$.

Since afterwards we intend to apply a clustering of the direction frames, losing all sequence information, we employ a windowing scheme to preserve temporal information.

\newcolumntype{C}[1]{>{\centering\let\newline\\\arraybackslash\hspace{0pt}}m{#1}}

\newcolumntype{D}{>{\centering\let\newline\\\arraybackslash\hspace{0pt}}m{0cm}}

\newcolumntype{x}[1]{>{\centering\let\newline\\\arraybackslash\hspace{0pt}}p{#1}}

\begin{figure}
\label{preprocessing_diagram}
\begin{tabular}{ C{0.95cm} C{0cm}  C{0.95cm} C{0cm}  C{0.95cm} C{0cm} C{0.95cm} } 

%\begin{tabular}{ x{0.95cm} x{0cm} x{0.95cm} x{0cm} x{0.95cm} x{0cm} x{0.95cm} } 

%\begin{tabular}{ c | c | c c c c c}
\includegraphics[height=40px]{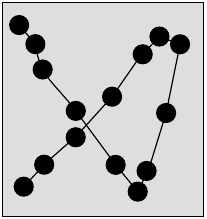} 
& \includegraphics[height=40px]{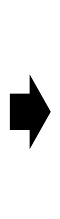}
&\includegraphics[height=40px]{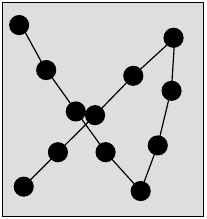}
&\includegraphics[height=40px]{figures/preprocessing_arrow}
&\includegraphics[height=40px]{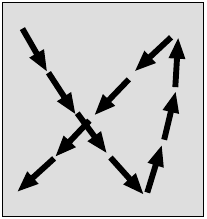}
&\includegraphics[height=40px]{figures/preprocessing_arrow}
&\includegraphics[height=40px]{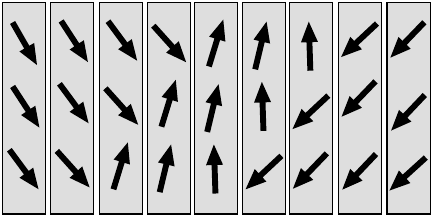}
\\ [-0.05cm]
\hspace{0.4cm} a &  & \hspace{0.4cm} b &  & \hspace{0.4cm} c &   & \hspace{0.6cm} d
\end{tabular}
\vspace{-0.3cm}
\caption{Preprocessing steps for a single joint of the action. The original positions (a) are smoothed and resampled obtaining a more uniform sequence (b). Then, direction vectors are computed (c) and concatenated into overlapping WDFs with $W=3$ (d).}
\end{figure}

From the $F-1$ direction frames, we generate $\fw=(F-1)-(W-1)=F-W$ \textit{windowed direction frames} (WDFs) by concatenating $W$ direction frames together in an overlapping fashion, such that:

\begin{equation}
\label{window} 
\wif = \left[ \dif, \difp, \dots, \difw \right], \quad f=1,\dots,\fw
\end{equation}

where $W$, the window size, is a parameter that represents how many direction frames are concatenated to form a WDF. As $W$ grows, more temporal information is retained. In an extreme case, if $W=F-1$, then $\fw=1$, and there is just one WDF which is exactly $\di$. On the other hand, if $W=1$, then $\fw= F-1$ and $\wif=\dif$, ie, the WDFs are simply the direction frames.

\subsection{Descriptor}

To calculate the descriptor, we first obtain the $K$ cluster centers $\cw = \{ \ci{1},\ci{2},\dots,\ci{K}\}$ of a clustering of the set of all WDFs of all actions $\wall = \{  \wif \, | \, i=1,\dots,N, f=1,\dots,F-W \}$ contained in the training set. The clustering $\cw$ is built with a Self Organizing Map (SOM) algorithm \cite{kohonen2001self}. We chose a SOM because of its simplicity and generally good behaviour, since we found no significant performance difference when employing other clustering algorithms.

Afterwards, we use the clustering $\cw$ to quantize the space of all possible WDFs, so as to compute the distribution of movements over that set. Therefore, given a  WDF $\wf$, we can assign it to a cluster with center $\cl$, according to the same distance metric employed in the generation of the cluster. 

To quantize the space, we calculate a normalized histogram $\hc$ with $K$ bins given by the cluster centers $\cl$, so that bin $l$ of the histogram, $\hl$, represents the normalized count of WDFs that belong to that cluster.

We denote the center of the clustering $\cw$ corresponding to $\wf$ as $BMU(\wf)$, ie, this is the center position of the cluster to which $\wf$ belongs. Then, remembering that the quantity $\fw$ represents the amount of WDFs generated for the action, $\hc$ is a histogram whose bins are $\hl$ as defined below:

\begin{equation}
\label{histogram}
\hl = \frac{ | \, \{ \wf \, | \, BMU(\wf) = \cl  \} \, | }{ \fw }, \quad l=1,\dots,K
\end{equation}

Where $|A|$ denotes the cardinality of set $A$. So $\hl$ is the percentage of WDFs generated for the action that belong to the cluster with center $\cl$.

The resulting descriptor, $\hc$, measures the distribution of WDFs of the original action $\ac$. The histogram elements $\hl$ can be interpreted as the frequentist estimation of the probability that an action $\ac$, transformed into $\hc$, contains WDF $\cl$ as a composing element, ie, $\hl = P( \cl  | \ac )$.

\subsection{Classification Model}
s

We have employed the classification model described in \cite{Estrebou2010} because it has given good results with a similar descriptor. Using the same training actions $\as$ we employed to build the clustering (the training set) we will build a model of $P(c | \ac)$, ie, the probability that an action $\ac$, is of class $c$. 

Since each element of $\hc$ gives information about $P( \cl  | \ac )$, the training set can be used to estimate those probabilities. Therefore, we will use the training set as well to build a model of $P(c | \cl)$ so we can factor $P(c | \ac)$ as:

\begin{equation}
\label{factorization}
P(c | \ac) = P(c | \cl) P( \cl | \ac)
\end{equation}

Since there are $K$ centers $\cl$ in our clustering, there are as many estimations of $P(c | \ac)$. Therefore, we can classify new actions by computing the probability of an action $\ac$ belonging to a certain class $c$ as the average of $P(c | \ac)$ over the $K$ cluster centers $\cl$:

\begin{equation}
\label{classactionmodel}
P(c | \ac) \propto \sum_{l=1}^{K} P(c | \cl) P( \cl | \ac) = \sum_{l=1}^{K} P(c | \cl) * \hl
\end{equation}

The estimation of $P(c | \cl)$ is performed as follows. Having built the cluster $\cw$, we are thus able to compute descriptors for the set of training actions $\as=\{ \ac^1,\dots,\ac^N \}$, for which we know the corresponding classes $c_i$. We can therefore build a model of $P(c | \cl)$ as:

\begin{equation}
\label{classwdfmodel}
P(c | \cl) = \frac{ | assigned(\cl) \cap ofClass(c) |}{|assigned(\cl)| }
\end{equation}

Where $\cap$ denotes set intersection, $ofClass(c)$ is the set of WDFs of class $c$, and $assigned(\cl)$ is the set of WDFs assigned to center $\cl$ (over all the training actions, in both cases). Formally:

\begin{align*}
\label{assigned}
assigned(\cl)= \{ \wif | BMU(\wif)=\cl, \\
\, i=1,\dots,N, \, f=1,\dots,\fw \}
\end{align*}

\begin{equation*}
\label{ofClass}
ofClass(c)= \{ \wif | c_i=c, \, f=1,\dots,\fw \}
\end{equation*}

Equation \ref{classwdfmodel} calculates $P(c | \cl)$ as the percentage of WDFs of the training set actions that were of class $c$ over those assigned to the center $\cl$.

The quantities $P(c | \cl)$ can be computed once based on the training set and used directly afterwards.

To assign a class label $c^{*}$ to an action $\ac$ we therefore compute the descriptor $\hc$ for the action, and then calculate:
\begin{equation}
c^{*} = \max_c P(c | \ac), \quad c=1,\dots,C
\end{equation}

where $C$ is the number of classes, $P(c | \ac)$ is defined as in eq. \ref{classactionmodel}, and $P(c|\cl)$ and $P(\cl | \ac)$ are estimated as in eqs. \ref{classwdfmodel} and \ref{histogram} respectively.

%% file: tex/experiments.tex
% !TeX spellcheck = en_US

\begin{figure}[!ht]
\centering
\includegraphics[scale=0.6]{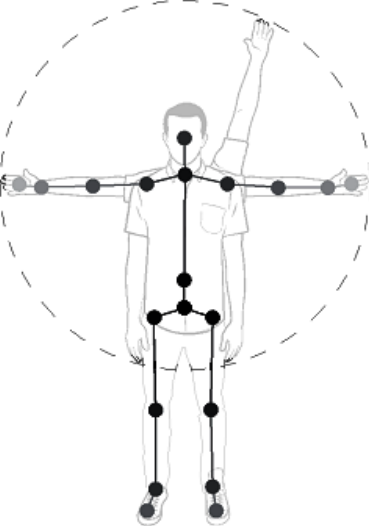}
\caption{Joints detected by Kinect}
\label{kinect_joints}
\end{figure}

We evaluated the descriptor on two well known and publicly available datasets, MSRC12 \cite{fothergill2012instructing} and MSR Action3D \cite{li2010action}, both acquired using a Kinect Sensor, which captures 20 joints of the human body (Figure \ref{kinect_joints}). \footnote{A detailed description of how we prepared the datasets and which action instances were corrected/left out, scripts to read the files, and code to run the experiments can be found in \url{http://facundoq.github.io/publications/dam/action_descriptor/}} 

For both datasets, we tested various values for $W$, the window size, to evaluate the importance of including temporal information in the descriptor and $K$, the number of clusters, to establish if the classifier was robust in a wide variety of settings of this parameter. The amount of frames the actions are resampled to, $F$, was set to 25 for all experiments with Action3D.  For the MSRC12 dataset, since the actions are easier to separate by means of joint activation patterns instead of trajectory information, the actions were resampled to $F=16$ frames without losing accuracy. After computing the WDFs, they were rescaled to euclidean norm $1$ so as to provide scale invariance to the descriptor.

\subsection{MSR Action3D Dataset}
\label{action3d}
\input{tex/action3d}

\subsection{MSRC12 Dataset}
\label{msrc12}
\input{tex/msrc12}

%% file: tex/action3d.tex
% !TeX spellcheck = en_US
\begin{figure*}[t]
\doublerulesep 0.1pt
\begin{footnotesize}
\begin{tabular}{ c c c } 
\includegraphics[height=50mm]{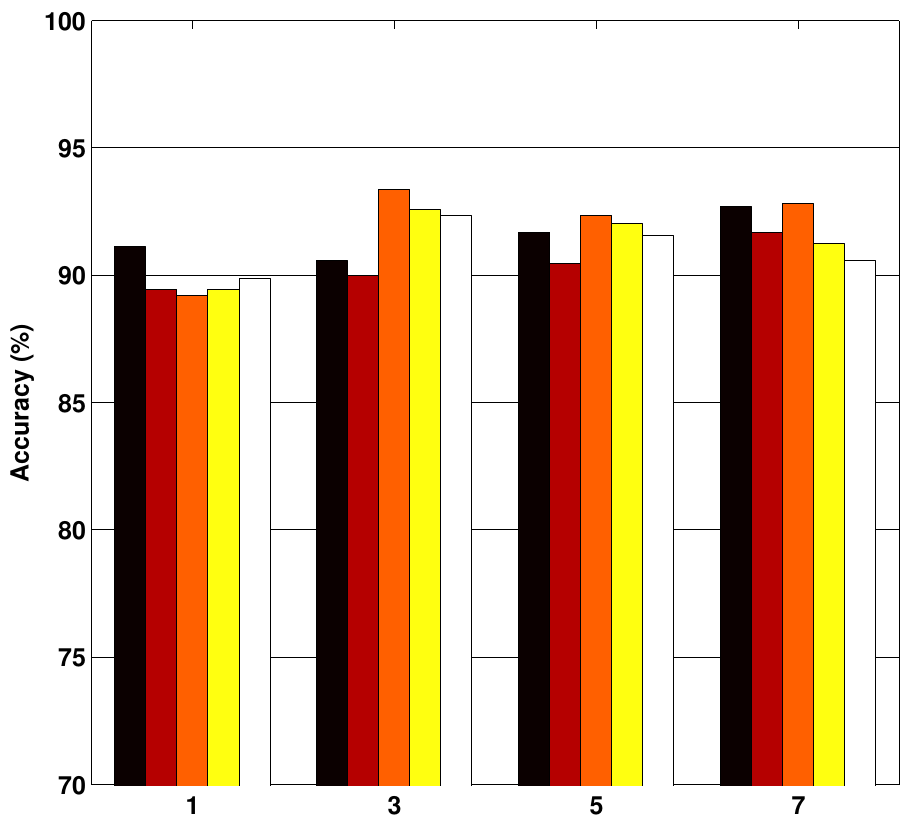} &
\includegraphics[height=50mm]{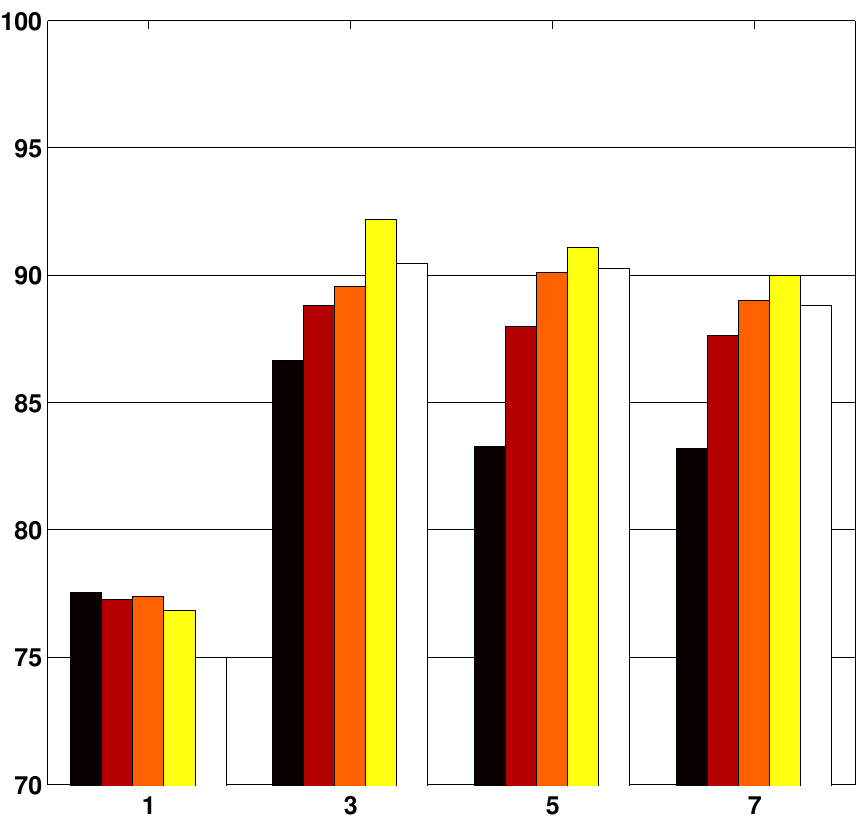} &
\includegraphics[height=50mm]{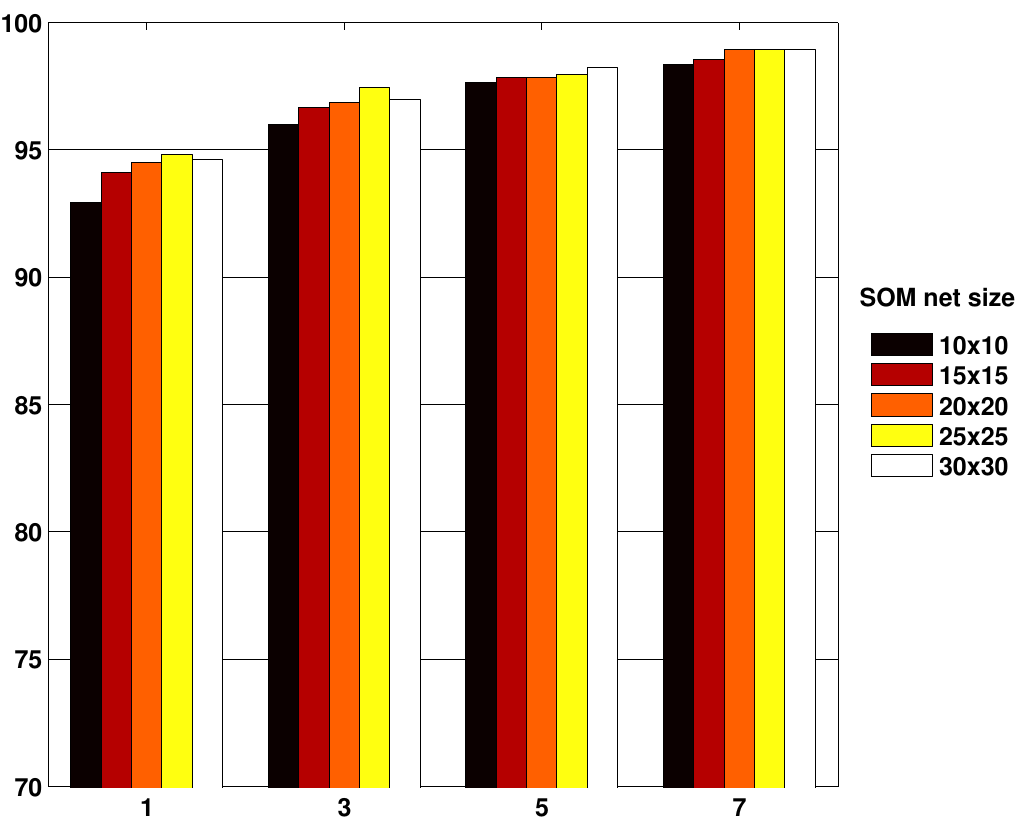} \\
AS1 & AS2 & AS3 \\
\end{tabular}
\centering
\wrappingcaption{Results obtained with our descriptor for MSR-Action3D, with different values for $W$, the window size, and $K$, the amount of clusters. Each result is the average of 30 independent runs.}
\label{figure_acc_Action3D}
\end{footnotesize}
\end{figure*}

We used the MSR Action3D dataset \cite{li2010action} which consists of action sequences of 20 different types of actions performed by 10 subjects. Each sequence consists of an action, segmented so that the start of the sequence roughly corresponds to that of the action, and the same for the end. For each sequence the dataset contains the position of the 20 joints for all the frames captured, which are time-labeled.

There are 567 sequences in total, of which we used 526. The original authors had already left out 10 sequences out of the 567; we excluded 31 more that we found heavily corrupted and corrected 48 in which some joints positions were exchanged\cite{li2010action}. 

\subsubsection{Experimental setup}
We employed the typical experimental setup on this dataset \cite{li2010action}, which divides the action classes into three action sets (AS1, AS2 and AS3), each composed of the actions of 8 classes, with some overlap between action sets.

%(see table \ref{action_sets})
We performed experiments on each set separately, following a cross subject protocol with a 50/50 subject split \cite{li2010action}. We performed 30 cross validation runs for each parameter configuration, and computed the average classification rate for each over the 30 runs. 

Our best results were obtained with a window size $W=3$ and $K=625$ clusters obtained after training a $25 \times 25$ SOM network.
% figura con resultados action3D (4 window, 4 SOM).  3 figuras --> AS1, AS2 y AS3

\subsubsection{Results}

% Acc Action3d comparativa
\begin{table}
\begin{footnotesize}

\wrappingcaption{Comparison of results on the MSR-Action 3D dataset.}
\begin{tabular}{ | p{5cm} | c | } 
	 \hline
	 \rowcolor{lightgray}
	 \centering \textbf{Method} & \textbf{ Accuracy (\%)}  \\ 
	 \hline
	 Ellis (logistic regression) \cite{Ellis2013} & 65.70 \\ 
	 \hline
 	 Li (Action Graph) \cite{li2010action} & 74.70 \\ 
	 \hline
	 Wang (Random Occupancy Pattern)  \cite{wang2012robust} & 86.50 \\ 
	 \hline
	 Wang (Actionlets Ensemble) \cite{wang2012actionlet} & 88.20 \\ 
	 \hline
	 Hussein (Cov3DJ) \cite{hussein2013human} & 90.53\\ 
	 \hline	 
	 \textbf{Proposed descriptor} & \textbf{94.07}  \\ 
	 \hline
\end{tabular}

\label{table_acc_action3D_comparativa}
\end{footnotesize}
\end{table}

Table \ref{table_acc_action3D_comparativa} compares the results obtained with the proposed method against other state-of-the-art models. Our descriptor achieves an accuracy of 94.07\%, reducing the error in $1/3$ with respect to the best method. 

%La tabla \ref{table_acc_action3D_comparativa} muestra los resultados obtenidos con el método propuesto comparados con los métodos más recientes en el estado del arte. Nuestro método logra un 94.07 \% de accuracy, reduciendo en más de $1/3$ el error del mejor método.

Figure \ref{figure_acc_Action3D} displays the different accuracies obtained by varying the parameters of the method for each subset of the dataset. It is evident that the effect of the windowing scheme is beneficial in almost all cases. On the other hand, the effect of the number of clusters in the model is mostly insignificant.

\begin{figure}
\includegraphics[width=0.4\textwidth]{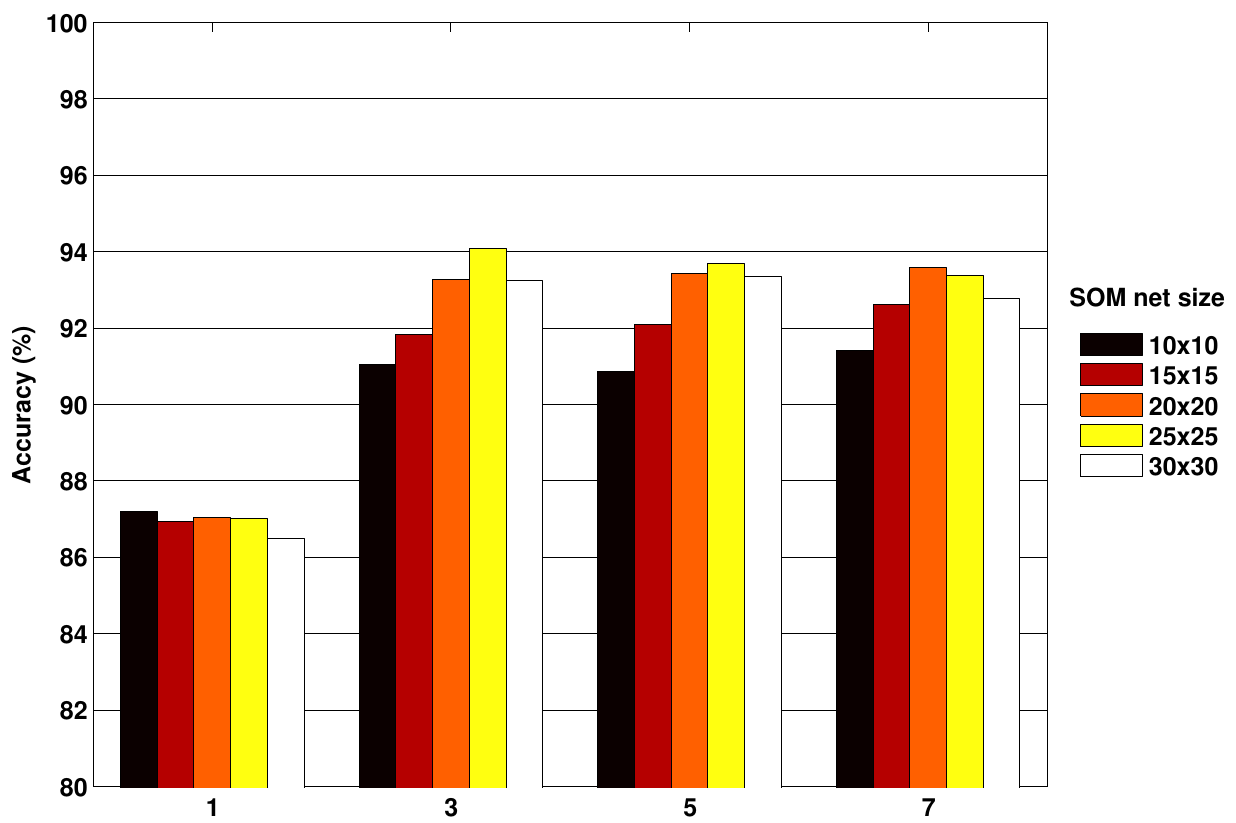}
\centering
\caption{ Average accuracy (\textit{y axis}) on MSR-Action3D for different window sizes $W$ (\textit{x axis}) and number of clusters, $K$ (colors). The average is computed over the accuracies of the 3 subsets (AS1, AS2, AS3).}
\label{figure_acc_AVG_Action3D}
\end{figure}

Figure \ref{figure_acc_AVG_Action3D} summarizes the results showing the average accuracy over the three datasets for all the parameter configurations evaluated . Lastly, table \ref{table_acc_action3D_SOM25} shows the results obtained in the each subset of the dataset separately with our best configuration  ($K=25x25$ neurons in the SOM network, $W=3$ window size). Note that results for our descriptor with $W=5$ and $W=7$ also outperform the second best known method listed in table \ref{table_acc_action3D_comparativa}.
 
%La figura \ref{figure_acc_Action3D} muestra el comportamiento del método propuesto al variar los parámetros establecidos. Cada gráfico representa los resultados de un dataset (AS1, AS2 y AS3). Es posible observar a simple vista el buen comportamiento de la ventana temporal, ya que al establecer la ventana en $1$ el descriptor es equivalente a tomar sólo un frame. Con excepción de AS2, los cambios en la cantidad de clusters utilizados no parecen ser significativos.
%La figura \ref{figure_acc_AVG_Action3D} resume los resultados mostrando el promedio de los 3 datasets, para todas las configuraciones evaluadas. Por último, la tabla \ref{table_acc_action3D_SOM25} muestra los resultados obtenidos en los 3 datasets junto con la media, utilizando una red SOM de $25x25$ neuronas (siendo esta la mejor configuración obtenida). El mejor valor obtenido para la media de AS1, AS2 y AS3 fue con $W=3$, aunque cabe destacar que $W=5$ y $W=7$ sobrepasan aún al segundo mejor método en la literatura (ver tabla \ref{table_acc_action3D_comparativa}).

 % figura con grafico del AVG action3D. 4 som 4 Wind

\begin{table}
\begin{footnotesize}
\wrappingcaption{Accuracy of the model on each subset of the MSR-Action 3D dataset, for different values of $W$ and a fixed number of clusters $K=25 \times 25$. Standard deviation shown in parenthesis.}
\begin{tabular}{ | c | c | c | c | c | } 
\hline
\rowcolor{lightgray}	
\textbf{Window}	& 1 & 3 & 5 & 7 \\
\hline
\textbf{AS1}	& 89.4 ($\pm$0.9) &	92.6 ($\pm$1.4)&	92.0 ($\pm$1.5)&	91.2 ($\pm$1.5)\\
\hline
\textbf{AS2}	& 76.8 ($\pm$1.9)&	92.2 ($\pm$1.6)&	91.1 ($\pm$1.3)&	90.7 ($\pm$1.0)\\
\hline
\textbf{AS3}	& 94.8 ($\pm$0.5)&	97.4 ($\pm$0.7)&	97.9 ($\pm$1.0)&	98.9 ($\pm$0.7)\\
\hline
\hline
\textbf{Mean} & \textbf{87.02} &	\textbf{94.07} &	\textbf{93.68} &	\textbf{93.61} \\
\hline
\end{tabular}
\label{table_acc_action3D_SOM25}
\end{footnotesize}
\end{table}

\begin{comment}
\doublerulesep 0.1pt
\begin{table}[h]
\begin{footnotesize}
\caption{Action sets for MSR Action3D} 
\label{action_sets}
\begin{tabular}{p{2cm}p{1cm}p{0.5cm}p{0.5cm}p{0.5cm}}
\hline\hline\noalign{\smallskip}
    Action & Sequences & AS1 & AS2 & AS3 \\
\noalign{\smallskip} \hline
    Tennis Swing & 20 & No & Yes & No\\
    Tennis Swing & 20 & No & Yes & No\\
    Tennis Swing & 20 & No & Yes & No\\
\hline\hline
\end{tabular}
\end{footnotesize}
\end{table}
\end{comment}

\subsubsection{Discussion}

%\textbf{HACER UN POCO DE PARLA SOBRE LAS VENTAJAS EN VELOCIDAD O SIMPLICIDAD}

%\textbf{(QUIZA DEBERIA SER UNA SOLO SECCION DISCUSSION PARA AMBOS DATASETS)}

In this section we perform a detailed analysis of the proposed method, exploring the results in each subset of the dataset separately.

%En esta sección hacemos un análisis más profundo del método desarrollado, observando los resultados obtenidos en cada subset.

Figure \ref{figure_acc_Action3D} shows a significant difference in accuracy between subsets. The method consistently reaches very high accuracies on Subset AS3, as high as $99\%$ in occasions. On the other hand, the method achieves much worse accuracies on AS2, in which most configurations are below $90\%$.

\begin{figure}
\includegraphics[width=0.45\textwidth]{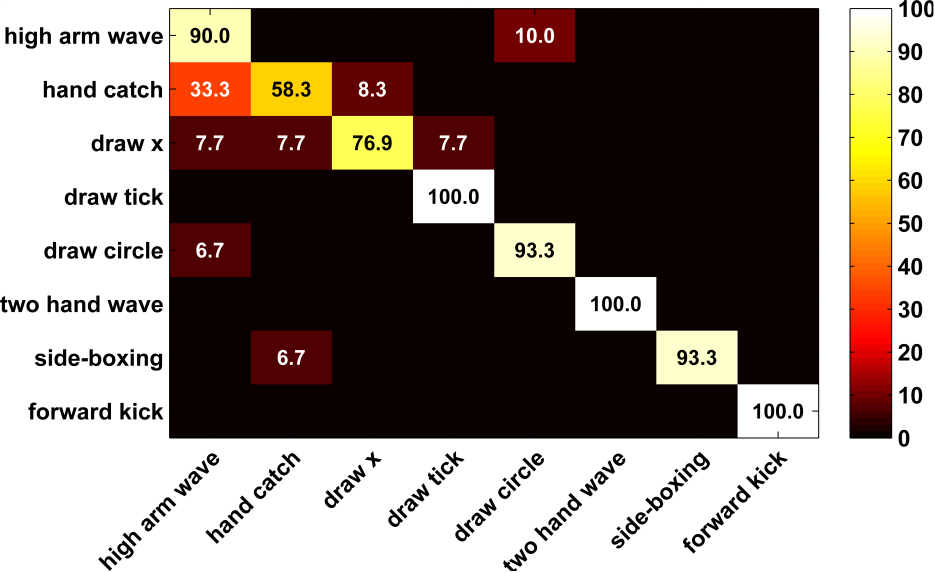}
\centering
\caption{Confusion matrix for subset AS2 of MSR-Action3D.}
\label{figure_confMatrix_AS2}
\end{figure}

Figure \ref{figure_confMatrix_AS2} shows the confusion matrix for a single run of the experiment on AS2, from which it can be deduced which actions affect the recognition rate the most. For example, actions  \textit{high arm wave} and \textit{hand catch} are performed with the same set of joints and are very similar in their motions, thus making our model less likely to discriminate between them. Other actions such as \textit{forward kick} or \textit{two hand wave} are naturally easier to distinguish since they involve the use of different joints altogether. This pattern in the difficulty of the dataset was already described in previous works \cite{wang2012actionlet}\cite{wang2012robust}.

%Si observamos la figura \ref{figure_acc_Action3D} es notable la diferencia en acuracy entre los 3 datasets. En el mejor caso, tenemos a AS3 con tasas de acierto muy cercanas al $100\%$. En el otro extremo, tenemos a AS2, quien no consigue una tasa de acierto mayor al $90\%$ para la mayoría de las configuraciones. 
%La figura \ref{figure_confMatrix_AS2} muestra la matriz de confusión para una ejecución particular en AS2 (el peor subset de los tres). Aquí es posible observar qué gestos son los que reducen la tasa de acierto en este subset. Por ejemplo, gestos como \textit{high arm wave} o \textit{hand catch} resultan muy similares entre sí, lo que hace muy similares a los descriptores. En cambio gestos como \textit{forward kick} o \textit{two hand wave} son naturalmente más sencillos de discriminar ya que utilizan diferentes joints. Algo similar es posible notar en artículos de la bibliografía \cite{wang2012actionlet}\cite{wang2012robust}.

 % Figura confusion Matrix AS2

 % Figura prob Matrix AS2
\begin{figure}
\includegraphics[width=0.45\textwidth]{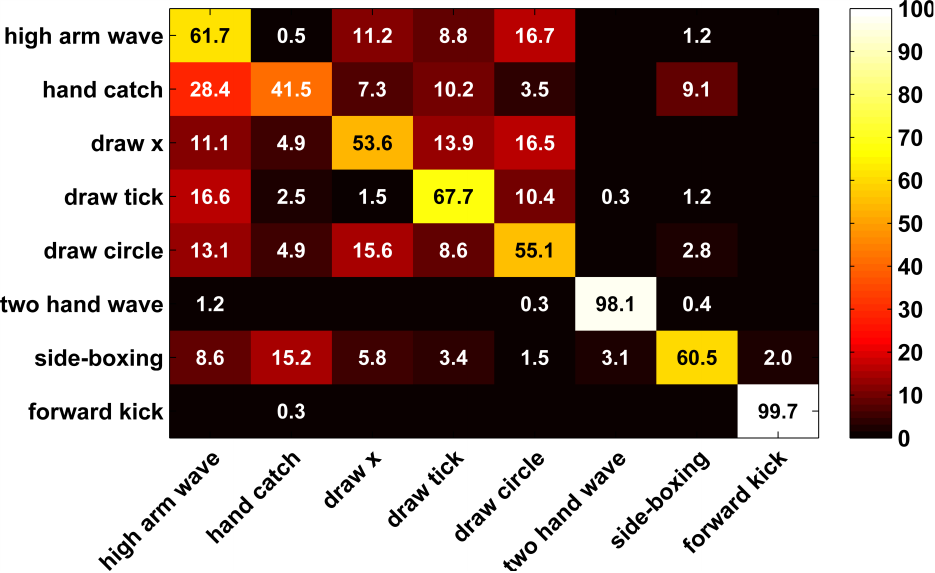}
\centering
\caption{ Average probability of assigning an action of class $c$ (rows) to every other class (columns) for subset AS2 of MSR-Action3D.}
\label{figure_probMatrix_AS2}
\end{figure} 

Figure \ref{figure_probMatrix_AS2} shows the probability matrix employed by the ProbSOM classification rule for the testing actions. The matrix shows the average estimated probabilities that an action of a certain class $C$ belongs to another class $C'$ (where possibly $C=C'$) as computed by our classification rule. It is remarkable that for some actions (\textit{forward kick},\textit{two hand wave}) not only the class prediction was perfect (\textit{forward kick},\textit{two hand wave}) but also that the probability of the action $\ai$ belonging to a class $c$ was $0$ for classes other than $c_i$ (the true class of the action $\ai$), ie, $P(c |\ai)=0$ if $c \neq c_i$. However, some actions which can be readily identified as similar from just their general trajectory did not have such a perfect probability assignment.

%Haciendo un análisis más profundo sobre la robustez del método propuesto, la figura \ref{figure_probMatrix_AS2} muestra la matriz de probabilidades utilizada por el ProbSOM para clasificar los gestos de testing. Dicha matriz refleja el promedio de las matrices de probabilidades para todos los gestos de cada clase. Es posible notar como algunos gestos no sólo fueron clasificados correctamente sino que la distribución de probabilidades fue perfecta (\textit{forward kick},\textit{two hand wave}). Por otro lado, también es posible notar la no tan óptima distribución que poseen algunos gestos que visualmente resultan muy similares.

%% file: tex/msrc12.tex
% !TeX spellcheck = en_US

The MSRC12 dataset is described in \cite{fothergill2012instructing} by Fothergill et al. and consists of 594 sequences of 12 different types of actions performed by 30 subjects. Each sequence contains the recorded positions of several instances or performances of the same action by the same person. The action types are described as: \textit{Crouch or hide, Shoot a pistol, Throw an object, Change weapon, Kick, Put on night vision goggles, Start Music/Raise Volume (of music), Navigate to next menu, Wind up the music, Take a Bow to end music session, Protest the music,} and \textit{Move up the tempo of the song}. The data was captured using a Kinect device. 

\subsubsection{Experimental setup}

Following the experimental setup by Hussein et al \cite{hussein2013human}, we segmented the sequences to isolate each action instance using their annotation, obtaining a set of 6244 action instances. 

By inspecting the actions visually we were able to detect instances in which the subject did not follow the instructions as intended by the experimenters. For example, subject 10 performs \textit{Kick} with his left leg, instead of the right one; for the action \textit{Wind it up}, subjects must trace a circle with both hands in front of them, but subject 9 makes it with only one hand (alternating right and left), and subject 21 traces only half of the circle. This was indeed expected by the experimenters, given that the database was generated precisely to measure how different instruction modalities (video, pictures, text) affect the performance of the actions, and so those actions performed incorrectly were not excluded from the dataset \cite{fothergill2012instructing}. The differences in execution are so wide that this outliers can be readily considered as instances of other types of action classes, and we have therefore excluded those instances from our experiments \footnote{The site https://sites.google.com/site/damdescriptor contains a full list of left out and corrected actions.}.

With the filtered dataset, we performed cross subject experiments with a 50/50 subject split, and 30 cross validation runs for each parameter configuration. We report the average classification rate for each configuration over the 30 runs. The window size for WDFs was set to $W=5$ and $K=900$ clusters where used in a $30 \times 30$ topography SOM network. We also performed leave-one-subject-out cross validation tests following the protocol described in \cite{hussein2013human}.

\subsubsection{Results}

% tabla con resultados para MSRC12 cross subject
\doublerulesep 0.1pt
\begin{table}
\begin{footnotesize}
\wrappingcaption{Comparison of results on the MSRC-12  
dataset with cross-subject validation.}
\begin{tabular}{ | p{6cm} | c | } 
	 \hline
	 \rowcolor{lightgray}
	 \centering \textbf{Method} &\textbf{ Acc. (\%)}  \\ 
	 \hline
 	 Ellis (Logistic reg.)\cite{Ellis2013} &  88.7\\ 
	 \hline
	 Hussein (Cov3DJ) \cite{hussein2013human} & 91.7 \\ 
	 \hline	 
	 \textbf{Proposed descriptor} & \textbf{91.7}  \\ 
	 \hline
\end{tabular}
\label{table_acc_MSRC-12_comparativa}
\end{footnotesize}

\end{table}

% tabla con resultados para MSRC12 Leave one out
\doublerulesep 0.1pt
\begin{table}
\begin{footnotesize}
\wrappingcaption{Comparison of results on the MSRC-12 dataset with leave-one-subject-out cross-validation. Note that in the case of Jiang et al's Hierarchical Model, a traditional leave-one-out cross-validation scheme was employed, training and testing with the same subject. }
\centering
\begin{tabular}{ | p{3.99cm} | p{1.9cm} | c | } 
	 \hline
	 \rowcolor{lightgray}
	 \centering \textbf{Method} &  \centering \textbf{Test Validation} & \textbf{ Acc. (\%)}  \\ 
	 \hline
	 Hussein et al. (Cov3DJ) \cite{hussein2013human}& leave-one-subject-out & 93.6 \\ 
	 \hline
	 Negin et al. (Decision Forest)\cite{negin2013decision}& leave-one-subject-out & 93.2 \\ 
	 \hline	 
	 Jiang et al. (Hierarchical Model) \cite{Jiang2013}& leave-one-out & 94.6 \\
	 \hline
	 \textbf{Proposed descriptor}& leave-one-subject-out & 93.1  \\ 
	 \hline
\end{tabular}
\label{table_acc_MSRC-12_leave-one-out}
\end{footnotesize}
\end{table}

%figura leave-one-out. acc por subject
\begin{figure}
\includegraphics[width=0.47\textwidth]{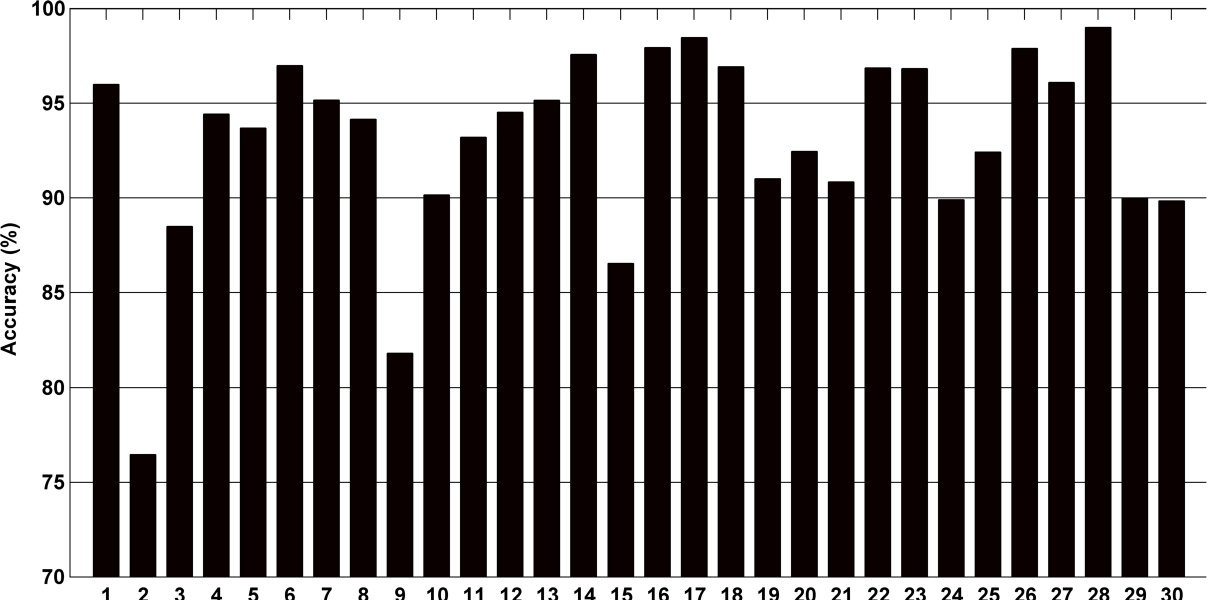}
\centering
\caption{Accuracy by subject on the MSRC-12 dataset for leave-one-subject-out cross-validation }
\label{figure_acc_MSRC12_subjects}
\end{figure}

%La tabla \ref{table_acc_MSRC-12_comparativa} muestra los resultados obtenidos para MSRC-12 con el método propuesto, comparado con los métodos más recientes en el estado del arte, utilizando un cross-subject validation method. Los reusltados son el promedio de 30 ejecuciones independientes. Si bien el Accuracy logró ser igual al método de Hussein\cite{hussein2013human}, cabe recordar el reducido tamaño del clasificador, lo que da la posibilidad de evaluarlo en tiempo real.

Table \ref{table_acc_MSRC-12_comparativa} shows the results obtained for MSRC12 with the proposed method, compared with the most recent ones of the state of the art. The experiments employ a cross-subject validation scheme, following \cite{hussein2013human}, but with subjects chosen randomly for each subject split. The results show the average accuracy on the dataset over 30 independent runs.

%La tabla \ref{table_acc_MSRC-12_leave-one-out} muestra los resultados obtenidos utilizando un leave-one-out validation method. Los reusltados son el promedio de 30 ejecuciones independientes. Este método de validación permite verificar la eficacia del método en un escenario clásico, donde un nuevo sujeto es insertado en el sistema.
%Jiang et al. \cite{Jiang2013} presenta un hierarchical model consiguiendo un accuracy de 94.6 \% (el mejor resultado encontrado en la bibliografía). Chatzis et al. utiliza un conditional random field-based model consiguiendo un 93\% de accuracy (***VERIFICAR***). Con nuestro método hemos logrado un accuracy de ??? \%, superando en un ?? \% al mejor método.

Table \ref{table_acc_MSRC-12_leave-one-out} shows results for a leave-one-subject-out validation scheme (with 30 runs), which allows to evaluate the descriptor's performance by simulating a classic scenario where a new subject tries to use a system trained by other people. In this setting, our method obtains an accuracy of 93\%, a similar result to those obtained by Negin et al., Hussein et al. and Jiang et al.

In both experimental settings, the variances for the average accuracy of our classification method were low ($\sigma^2=5$), but standard hypothesis tests for the differences in accuracies between classifiers could not be performed because neither the variances for the classification accuracy nor the code for performing the tests was available for any of the other methods considered. Nonetheless, the variances obtained seem appropriate given that the test results depend highly on the subjects chosen for testing, as explained in the next paragraph.

%Jiang et al's \cite{Jiang2013} hierarchical model obtains an accuracy of 94.6\%, but the results are not truly comparable as Jiang employs a simpler leave-one-out cross validation approach, which consists of a task considerably less difficult since samples of the same subject are used both for training and testing. 

Figure \ref{figure_acc_MSRC12_subjects} shows the accuracy obtained for each subject separately in the leave-one-subject-out cross validation experiments. The graph shows that there is a high amount of variation in the accuracies for each subject, which can be attributed to the different methods the subjects were instructed with (video, pictures, text). For example, subject 28 achieves an accuracy of nearly 100\%, while on the other hand for subject 2 it is only 76\%. This data suggests that the observed differences in classification accuracy between our descriptor and other models should not be statistically significant.

%La figura \ref{figure_acc_MSRC12_subjects} muestra el Acc obtenido para cada sujeto en particular, entranando con el resto y dejando a este como muestras de testing. El gráfico deja ver la notable diferencia que otorga el testing con cada uno de los sujetos del dataset. Por ejemplo, el sujeto 28 otorga un Acc de casi 100\%, mientras que en el otro extremo el sujeto 2 tiene un Acc de sólo 76\%. Esto se debe principalmente a los diferentes modos en que cada sujeto realiza los gestos. Cabe recordar que en MSRC12 dataset cada individuo esta instruido de diferentes modos, \textbf{SE PODRA COMPRABAR SI EL SUJETO 2 ESTA MAL INSTRUIDO??} donde algunos sujetos sólo tuvieron un texto describiendo el gesto, mientras que otros obtenían un video demostrativo.

 % figura con grafico del AVG MSRC12. 3 som 4 Wind
\begin{figure}
\includegraphics[width=0.48\textwidth]{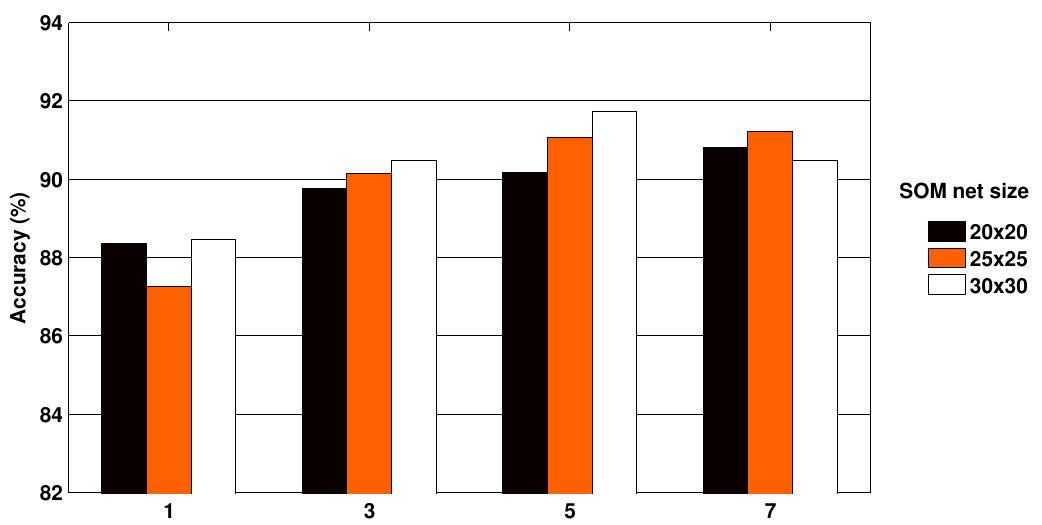}
\centering
\caption{Accuracy on the MSRC-12 dataset for different window and cluster sizes. $x-axis =$ window sizes, $y-axis$ = Method Accuracy}
\label{figure_acc_MSRC12_param}
\end{figure} 

Figure \ref{figure_acc_MSRC12_param} shows the behaviour of the method for different window and cluster sizes. As with MSRAction3D (subsection \ref{action3d}), the bar graphs clearly show the advantage of using WDFs instead of simple direction frames. On the other hand, the difference in accuracy from varying the number of clusters is much less significant. Given that the success of the ProbSOM classification rule is ultimately determined by how well the estimation of the distribution of WDF for each class was performed, and that estimation on the diversity in performing a certain action, the amount of clusters actually depends highly on the amount of samples in the dataset.

% \textbf{(I don't get this: if the amount of samples is the same across the dataset, what does it have to do with the fact that varying the number of clusters changes the accuracy? and with the fact that ProbSom is a statistical method? I'd just say that the network size is bigger for MSRC12 simply because there are more subjects and instances per subject, and therefore there is more intra-class variation to model.)}

%La figura \ref{figure_acc_MSRC12_param} muestra el comportamiento del método utilizando diferentes valores para la ventana temporal y la cantidad de clusters. Al igual que lo ocurrido con MSR Action3D (ver sección \ref{action3d}**ARREGLAR NUM**) puede observarse el comportamiento de la ventana temporal, ya que $W=1$ hace decaer el acuracy rápidamente. Las diferencias en la cantidad de clusters utilizados no es realmente significativa, y ya que el ProbSOM es un método estadístico, la cantidad de cluster depende también de la cantidad de muestras en el dataset.

\subsubsection{Discussion}  

 % Figura confusion Matrix MSRC12
 % Figura Prob Matrix MSRC12
\begin{figure}
\includegraphics[width=0.48\textwidth]{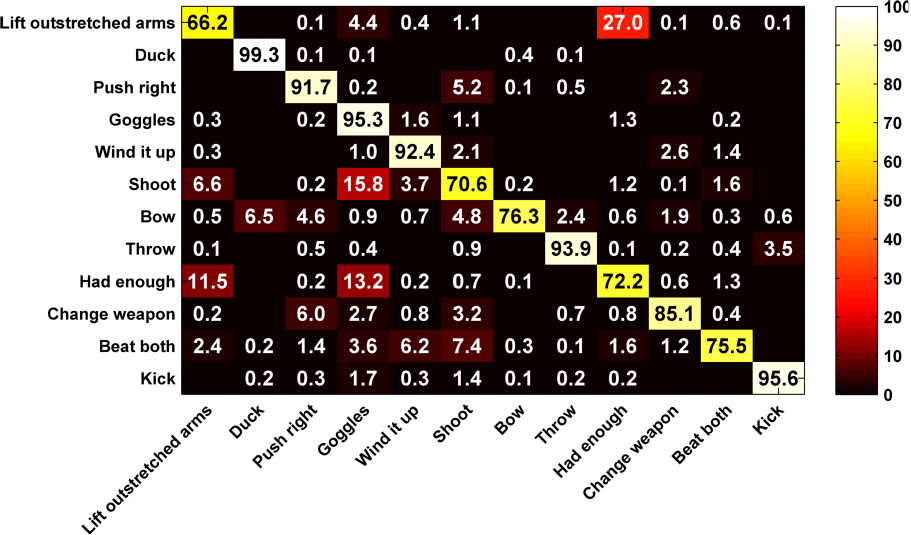}
\centering
\caption{Average probability of assigning an action of class $c$ to every other class for MSRC-12 dataset with leave-one-subject-out validation.}
\label{figure_confMatrix_MSRC12_prob}
\end{figure} 

\begin{figure}
\includegraphics[width=0.48\textwidth]{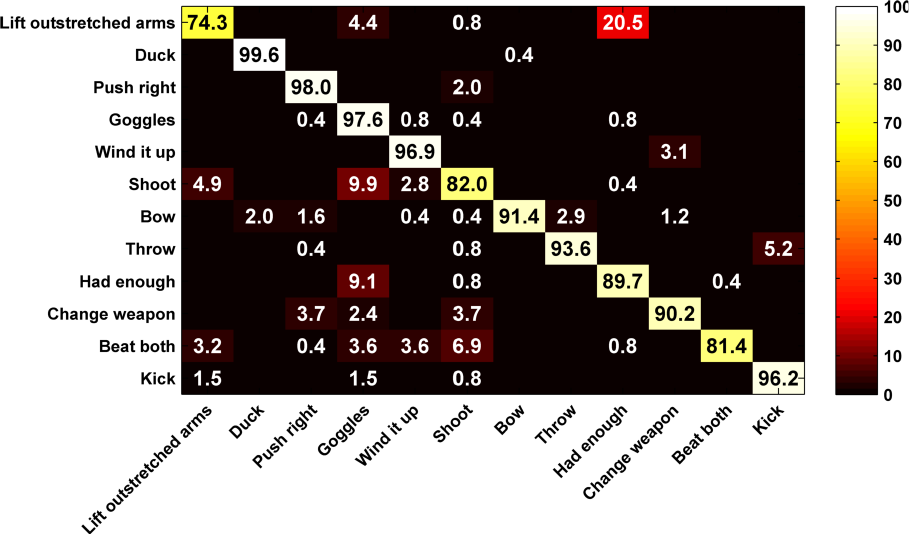}
\centering
\caption{Confusion matrix for MSRC-12 dataset with leave-one-subject-out validation. }
\label{figure_confMatrix_MSRC12}
\end{figure}

The method compares favorably with other state-of-the-art classification techniques for the MSRC12 dataset using traditional cross-subject validation. 

Although the accuracy our method achieves is the same as Hussein's \cite{hussein2013human}, their descriptor has a very high dimensionality; the best results they obtain are with a descriptor with $L=2$ overlapping levels, for which a skeleton represented with 20 joints results in a descriptor of length 7320 \cite{hussein2013human}.  Our descriptor is of length at most 900 in the case of a $30\times30$ SOM network. Coupled with the fact that the computation of the descriptor involves only simple arithmetic operations, it should provide better real-time performance in terms of efficiency and frame rate.

The method also shows state-of-the art performance when tested with a leave-one-subject-out cross-validation scheme, making it clear that new users can expect a robust response when using a human action recognition system that employs the descriptor.

%Al evaluar el método propuesto con MSRC12 dataset, se ven resultados comparables con los métodos más recientes en la bibliografía utilizando cross-subject validation. El método mostró mejores resultados al utilizar un leave-one-out cross-validation method, dando robustez a la posibilidad de incorporar nuevos gestos al clasificador.

As with the Action3D dataset, figure \ref{figure_confMatrix_MSRC12_prob} shows the distribution of probabilities used by the method to classify actions in the MSRC12 dataset. This is the average of the probabilities for all actions, with the average for each class computed with all the test actions of the class. We also present a confusion matrix for the same dataset and testing scheme in figure \ref{figure_confMatrix_MSRC12}. 

Some actions(i.e \textit{Duck} or \textit{Push Right}) were classified almost perfectly, with a low entropy distribution, which presents a strong case for the robustness of the model. Certain actions, like \textit{Lift outstretched arms} or \textit{Shoot} present more difficulties to the descriptor, presumably because they employ similar joints and direction movements. However, the confusion matrix in figure \ref{figure_confMatrix_MSRC12} also shows that the misclassification errors mostly involve two or three classes at a time, and so a second layer of classifiers could be trained to separate those specific classes which when together raise the classification error.

%Al igual que lo analizado para Action3d dataset, la figura \ref{figure_confMatrix_MSRC12} muestra la distribución de probabilidades utilizada por el método para clasificar los patrones en MSRC12 dataset\footnote{la matriz muestra el promedio de las distribuciones de todos los gestos para cada clase.}. Por su parte, la figura \ref{figure_confMatrix_MSRC12} muestra la confusion matrix para el mismo dataset. Puede notarse que ciertos gestos (i.e \textit{kick} or \textit{Duck}) fueron clasificados perfectamente, e incluso, cabe destacar que su distribución de probabilidades resultaron casi óptimas, no confundiéndose que ninguna otra clase. Esto le da una gran robustez al modelo.  Ciertos gestos, como \textit{Lift outstretched arms} o \textit{Shoot} resultan más confusos para el descriptor, confundiéndose con otras clases. No obstante, viendo la matriz de la figura \ref{figure_confMatrix_MSRC12} una ventaja del método propuesto radica en que la confusión suele ser con sólo una clase. Esto permite reducir el problema considerablemente ya que luego se podría hacer una nueva revisión haciendo énfasis en esas dos o tres clases.

%% file: tex/conclusion.tex
% !TeX spellcheck = en_US

We have defined a descriptor for human action recognition based on the distribution of action movements (DAM descriptor). 

The descriptor is computed as a histogram over a set of canonical or representative movements windowed directions, which are in turn obtained by performing a clustering with a SOM network,  over the set of all windowed directions in the training set. We believe that a key step in the generation of the descriptor is the windowing scheme that, unlike other authors' approaches, is included as an intermediate step and not with a pyramidal scheme.

The descriptor is both scale and translation invariant.  If desired,  direction invariance of the trajectory of an action can be achieved by inverting the direction of each joint in the direction frames. Rotation invariance can be incorporated as well by rotating the user to a canonical rotation using his position and pose to set the axes of the joint positions instead of the camera's. 

The descriptor shows state-of-the-art performance on standard cross subjects tests on two well known datasets: MSRC12 and MSRAction3D achieving a classification rate of 91.7\% and 94\% respectively. Additional leave-one-subject-out tests on MSRC12 have also shown state-of-the-art results with an accuracy of 93\%. 

Our future work will concentrate on including more datasets in our tests to evaluate the generalization performance of the method with other classes, further testing with pyramidal schemes to better include temporal information about the action, and evaluating its performance in online classification tasks.

%% file: tex/biographies.tex
% !TeX spellcheck = en_US
%% \Biography{#1}{#2}
%% #1 is the file name of photo,
%% #2 is author's biography

\Biography{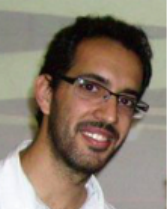}{Franco RONCHETTI is an advanced PhD student in Computer Science at the School of computer Science of the UNLP. Teaching Assistant for Grade Courses at the National University of La Plata. Between 2008-2011, recipient of a scholarship granted by the Ministry of Education of the Nation for the strengthening of human resources in the area of Information and Communication Technologies. Carrying out research activities since 2010 at the Institute of Research in Computer Science III-LIDI in areas related to soft-computing and intelligent systems, with focus on dynamic problems such as signal processing and pattern recognition, with various publications in the field.}

\Biography{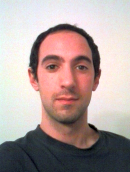}{Facundo QUIROGA received his BS in Computer Science in 2014 from the Faculty of Informatics of UNLP, Argentina. He is currently a PhD Student in the same Faculty, doing research work at the III-LIDI Institute. His main research fields comprise machine learning and signal processing, with applications in action recognition and biomedical data analysis. He is also interested in computational neuroscience and the interplay between artificial neural networks and biological ones.
}
\pagebreak

\Biography{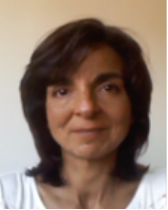}{Laura LANZARINI is a full-Time Head Professor since 2001 at the School of Computer Science of the UNLP, teaching in Data Mining topics. Guest Head Professor at the National University of Tierra del Fuego. Guest Professor of the Post-Grade Program at the University of Buenos Aires. Distinguished by the UNLP in December 2010 with the Award to Scientific, Technological, and Artistic Work for Senior Researchers. Director of the group "Intelligent Systems," part of the Institute of Research in Computer Science III-LIDI. Author of numerous publications in journals, congress proceedings and scientific workshops with international referees in subjects pertaining to the design and development of adaptive application based on neural networks, swarm intelligence, and other metaheuristics applicable to Data Mining and Text Mining problems.}

\Biography{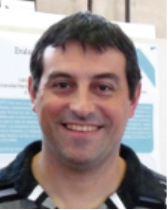}{Cesar ESTREBOU is a PhD student in Computer Vision, Image Processing and Graphic Computing at the School of Computer Science of the UNLP. Since 2008 is Teaching Assistant (Assignments Supervisor) for Grade Courses at the National University of La Plata. He has been carrying out research activities since 2005 at the Institute of Research in Computer Science III-LIDI in areas related to intelligent systems, real-time systems and software engenieering, with publications in journals, congress proceedings and scientific workshops with international referees in those fields.}